\documentclass[letterpaper, 10 pt, conference]{ieeeconf}  % Comment this line out
\IEEEoverridecommandlockouts                              % This command is only
\usepackage{xcolor}
\usepackage{amssymb}
\usepackage{latexsym}
\usepackage{float}
\usepackage{amsmath}

\usepackage{amsthm}
\usepackage{amsfonts}
\usepackage{mathrsfs}
\usepackage{array}
\usepackage{mathtools}
\usepackage{tikz}
\usepackage[demo,rel]{overpic}
\usepackage{rotating}
\usepackage{graphicx}
\usepackage{color}
\usepackage{multirow}
\usepackage[normalem]{ulem}
\usepackage{siunitx}
\usepackage{xcolor}
\usepackage{footnote}
\usepackage{pdflscape}
\usepackage{tablefootnote}
\usepackage[perpage]{footmisc}
\usepackage{todonotes}
\usepackage{caption}
\usepackage[list=on,listformat=simple]{subcaption}
\usepackage{epsfig}
\usepackage{tabularx}
\usepackage{hyperref} % must have ``breaklinks'' option enabled
\hypersetup{
  hypertexnames= false,
  breaklinks   = true,
  plainpages   = true,
  hidelinks    = true,
  colorlinks   = true, %Colours links instead of ugly boxes
  urlcolor     = blue, %Colour for external hyperlinks
  linkcolor    = black, %Colour of internal links
  citecolor    = black %Colour of citations
}

\usepackage{cite}
\usepackage{times}
\usepackage{algorithm}
\usepackage{algorithmicx}
\usepackage{algpseudocode}
\floatname{algorithm}{Procedure}

\usepackage{multicol}
\usepackage{wrapfig}

\title{\LARGE \bf
Rapid Trajectory Optimization Using C-FROST \protect\\ with Illustration on a Cassie-Series Dynamic Walking Biped}

\author{Ayonga Hereid$^{1}$, Omar Harib$^{2}$, Ross Hartley$^{2}$, Yukai Gong$^{2}$, and Jessy W. Grizzle$^{2}$% <-this % stops a space
\thanks{$^{*}$Ayonga Hereid and Omar Harib contributed equally to this work.}%
\thanks{$^{1}$Ayonga Hereid is with the Mechanical and Aerospace Engineering, Ohio State University, Columbus, OH 43210 USA {\tt\small hereid.1@osu.edu.}}%
\thanks{$^{2}$The authors are with the College of Engineering and the Robotics Institute, University of Michigan, Ann Arbor, MI 48109 USA {\tt\small \{oharib, rosshart, ykgong, grizzle\}}@umich.edu.}%
}
\usepackage{preambles}
\begin{document}
% \mainmatter              % start of a contribution  
% \title{\LARGE \bf
% Rapid Bipedal Gait Design Using C-FROST \\ with Illustration on a Cassie-series Robot
% }
% \titlerunning{Rapid Bipedal Gait Design Using C-FROST}  % abbreviated title (for running head)

% \author{Ayonga Hereid\thanks{Ayonga Hereid and Omar Harib contributed equally to this work.}, Omar Harib$^*$, Ross Hartley, Yukai Gong, and Jessy W. Grizzle}% <-this % stops a space
% \institute{University of Michigan, Ann Arbor MI 48109, USA}
% \authorrunning{Hereid \& Harib et al.} % abbreviated author list (for running head)

\maketitle
\thispagestyle{empty}
\pagestyle{empty}

\begin{abstract}
  One of the big attractions of low-dimensional models for gait design has been the ability to compute solutions rapidly, whereas one of their drawbacks has been the difficulty in mapping the solutions back to the target robot.  This paper presents a set of tools for rapidly determining solutions for ``humanoids'' without removing or lumping degrees of freedom.  The main tools are: (1) C-FROST, an open-source C++ interface for FROST, a direct collocation optimization tool; and (2) multi-threading.  The results will be illustrated on a 20-DoF floating-base model for a Cassie-series bipedal robot through numerical calculations and physical experiments.
\end{abstract}

\section{Introduction}
\label{sec:intro}

% TODO:: - Introduce optimizations - Mention Drake - Cite papers where we used fmincon - Cite direct collocation ayonga - Cite paper with Frost

% Motivations: - The need of use full body dynamics for gait/trajectory optimization (better algorithms) - The need of fast optimization (reduce computation overheads, remove Coriolis terms) - The need to generate a large number of gaits for SML/Mechanical Design process (parallel computation) - look at CSM Rabbit paper for the design process

Low-dimensional inverted pendulum models are by far the most common approach in the literature to getting around the analytic and computational challenges imposed by today's humanoid robots \cite{Kajita20013D, Pratt2012Capturability}.  In this approach, one maps a locomotion task onto the motion of the center of mass of a linear inverted pendulum (LIP), a spring-loaded inverted pendulum (SLIP), or other simplified models \cite{McGeer1990Passive} and equips the model with a ``foot-placement strategy'' for stability \cite{KAYAKO92, Raibert1986Legged, Pratt2006Velocity}. The overall simplicity of this approach is one of its uncontested advantages. On the other hand, when designing agile motions, guaranteeing stability in the full-order model and exploiting the full capability of the machine, especially in light of physical constraints of the hardware or environment, are among its main drawbacks.

\begin{figure}[tb]
  \centering
  \begin{subfigure}[t]{0.25\columnwidth}
    \centering \includegraphics[width=1\columnwidth, keepaspectratio]{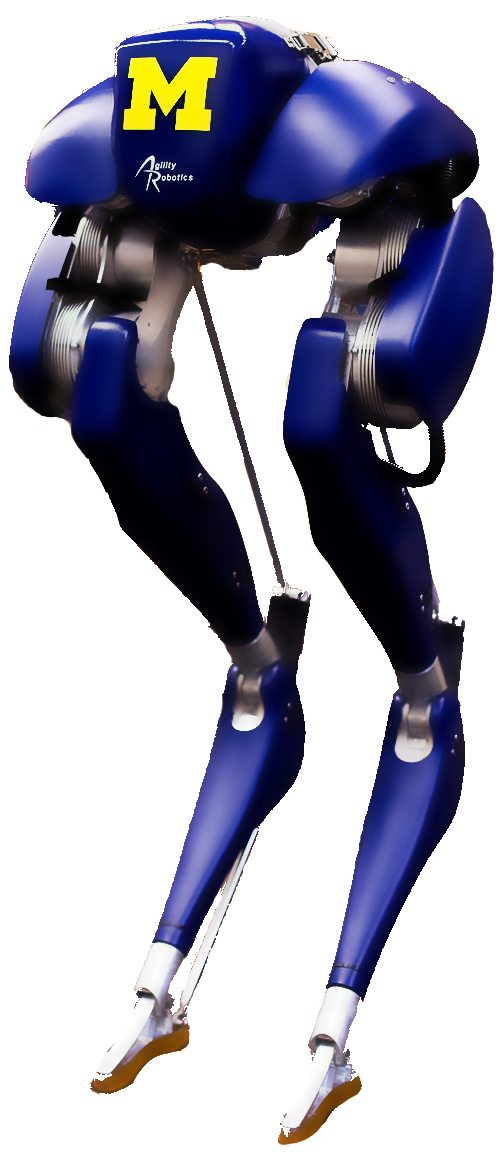}
    \caption{Cassie Blue}
    \label{fig:cassie:image}
  \end{subfigure}
  \begin{subfigure}[t]{0.7\columnwidth}
    \centering \includegraphics[width=1\columnwidth, keepaspectratio]{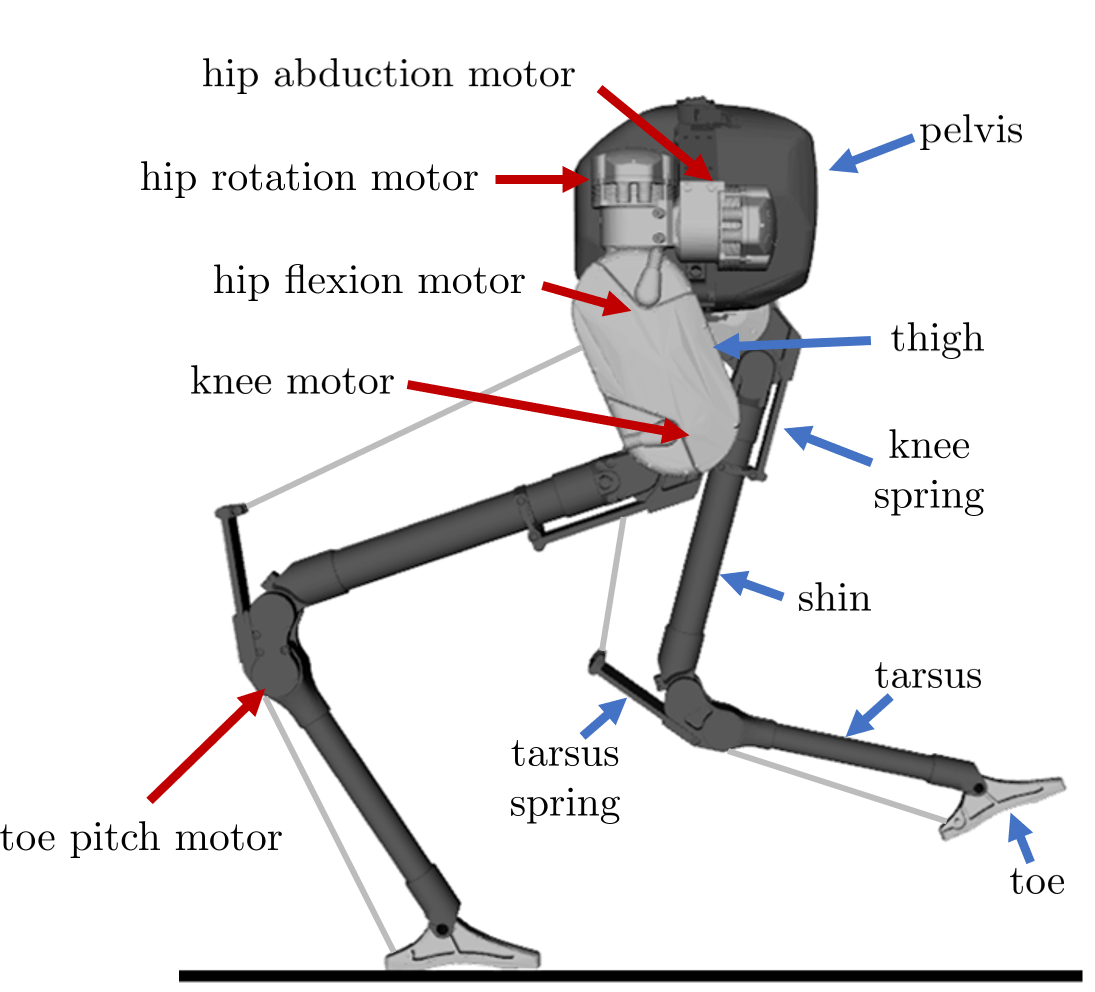}
    \caption{Cassie's Coordinates}
    \label{fig:cassie:coordinates}
  \end{subfigure}
  \caption{(\subref{fig:cassie:image}) Shows Cassie Blue, one of the Michigan copies of the Cassie series of robots built by Agility Robotics. (\subref{fig:cassie:coordinates}) Shows the joints and the kinematic model of the robot.}
  \label{fig:cassie}
\end{figure}

Because of these drawbacks, numerous groups are pursuing systematic approaches to utilize a high-dimensional mathematical model of a legged robot when designing dynamic behaviors and also when designing controllers that are sufficiently robust to realize these behaviors on the physical hardware. Speaking first to dynamic gait design, there is an increasing trend in the development of optimization-based planning algorithms which directly utilize more complicated dynamic models \cite{Carpentier2016Versatile, Dai2014Whole, Herzog2015Trajectory, Kuindersma2016Optimization}. These applications typically use general-purpose nonlinear programming solvers, e.g., IPOPT~\cite{Waechter2005implementation} or SNOPT~\cite{Gill2005SNOPT}, to formulate the motion planning problems. Others are advocating state-of-the-art optimal control toolboxes, such as GPOPS~\cite{Patterson2014GPOPS}, DIRCOL~\cite{Stryk1999Users} or PSOPT~\cite{Becerra2010Solving}, that come with advanced trajectory optimization algorithms. These latter packages, however, either lack an effective framework for users to construct the robot model or are not capable of solving large-scale problems that are very common for today's agile walking robots.

More recently, several open-source robotic toolboxes, including DRAKE \cite{drake} and FROST \cite{Hereid2017FROST}, have been developed to provide an integrated modeling, planning, and simulation environment for complex robotic systems. Both DRAKE and FROST are capable of optimizing dynamic walking gaits for high degree of freedom (DOF) robots using the full-order natural dynamic models of the machines. While DRAKE is implemented in a C++ environment for efficiency, FROST is developed as a MATLAB package for rapid prototyping. Compared to DRAKE, which uses automatic differentiation for computing gradient information for the NLP solvers, FROST takes advantages of the symbolic computation to generate the sparsity structure of the trajectory optimization problem. To speed up the computational speed for high-dimensional robotic systems, FROST is equipped with a custom symbolic math toolbox based on the Wolfram Mathematica kernel as the back-end to generate optimized C++ code for computing system dynamics and kinematics functions. Moreover, FROST is capable of computing the analytic derivatives of optimization constraints and cost functions symbolically, and \emph{even more importantly, it can precisely determine the sparsity structure of the nonlinear trajectory optimization problem}. With these features, FROST generates dynamic biped walking gaits using the full-order dynamics for a full-size humanoid robot (e.g., DURUS) in less than 10 minutes \cite{Hereid2016Thesis}.

Since its inception, FROST has been used for fast dynamic gait generation of various bipedal robots \cite{Harib2018Feedback, CassieYouTubePlayList, Reher2016Realizing}. Yet, its performance is significantly affected by the computational overhead introduced by MATLAB. This becomes evident particularly in the case when many different gait trajectories are required to be designed \cite{da2017combining}.

% While FROST improves the convergence speed of gait optimization significantly when compared to traditional shooting approaches \cite{Westervelt2007Feedback}

% \begin{figure}[tb]
%   \vspace*{-1cm} \centering \includegraphics[height=8cm]{res/cassie.png} \vspace*{-0.4cm}
%   \caption{Cassie Blue, one of the Michigan copies of the Cassie series of robots built by Agility Robotics.}
%   \label{fig:cassie}
% \end{figure}

In this paper, we present C-FROST, an open-source package that converts the direct collocation trajectory optimization problem constructed in FROST to a more computationally efficient C++ code. By reducing the computational overhead of MATLAB functions, we are able to increase the speed over previous full-order dynamic optimization problems by a factor of six to eight for a single gait. The implementation of C-FROST also makes it easier to deploy parallel computation on cloud servers for the generation of a large set of different gaits, a feature that allows the gait library approach presented in \cite{da2017combining} to be practical on high-dimensional 3D bipedal robots and advanced exoskeletons \cite{Harib2018Feedback}. Beyond generating families of gaits for various walking speeds, directions, and terrain slopes, the ability to generate gaits in parallel for a family of models is also useful for iterative design of legged mechanisms \cite{Chevallereau2003RABBIT, Reher2016Realizing}.
% Finally, as is well known for manipulators, we show that for bipedal robots, in many instances, the ``velocity terms'' in the model can be neglected, and doing so significantly speeds up the optimization process.

The remainder of the paper is organized as follows. \secref{sec:frost} overviews the FROST package and the optimization problem being solved here for gait design. \secref{sec:cfrost} introduces a stand-alone open-source C++ package that carries out the optimization problem formulated in FROST. \secref{sec:results} presents a case study of the application of C-FROST for a Cassie robot, and its execution speeds are illustrated in \secref{sec:experiments} in terms of:
\begin{itemize}
\item optimization in FROST vs. C-FROST,
\item serial execution vs. parallel, and
\item desktop vs. the cloud.
\end{itemize}
\secref{sec:experiments} also demonstrates that the gait designs performed in the paper are viable on a Cassie robot. Conclusions are given in \secref{sec:conclusion}.

%%% Local Variables:
%%% mode: latex
%%% TeX-master: "main"
%%% End:

% \input{section_testbed.tex}
\section{Trajectory Optimization in FROST}
\label{sec:frost}

% TODO: Briefly introduce FROST and direct collocation
FROST (Fast Robot Optimization and Simulation Toolkit) is an open source MATLAB toolbox that provides an integrated development framework for mathematical modeling, trajectory optimization, model-based control design, and simulation of complex robotic systems. While initially developed particularly for bipedal robots, FROST has been applied to a wide range of robotic systems, including quadrupeds \cite{hamed2018dynamically}, robot manipulators \cite{Kolathaya2018Direct}, and the adaptive lane keeping control design of an automated truck model \cite{chen2017enhancing}. The most important feature of FROST is the fast trajectory optimization algorithm for high-dimensional hybrid dynamical systems using direct collocation approaches \cite{Hereid2017FROST}.

% In this section, we briefly review the trajectory optimization problem formulation in FROST.

\subsection{Robot Behavior Modeling}

In FROST, a particular behavior of a robot can be modeled as a hybrid system that exhibits both continuous and discrete dynamics. For instance, legged locomotion often consists of a collection of continuous phases, with discrete events triggering transitions between neighboring phases \cite{Grizzle2014Models}. 
\begin{definition}
  \label{defn:hybrid-control-system}
  A \emph{hybrid system} is a tuple,
  \begin{align}
    \label{eq:hybrid-control-system}
    \HybridControlSystem = (\DirectedGraph,\Domain,\ControlInput,
    \Guard,\ResetMap,\emph{FG}),
  \end{align}
  where $\DirectedGraph = \{\Vertex,\Edge\}$ is a \emph{directed graph}, $\Domain$ is the admissible configuration of the continuous domains, $\ControlInput$ represents the admissible controllers, $\Guard$ determines guard conditions that trigger discrete transitions, and $\ResetMap$ and $\emph{FG}$ represent the discrete and continuous dynamics respectively \cite{Hereid2017FROST}.
\end{definition}

\begin{figure*}
  \vspace{2mm} \centering \includegraphics[width=0.85\textwidth]{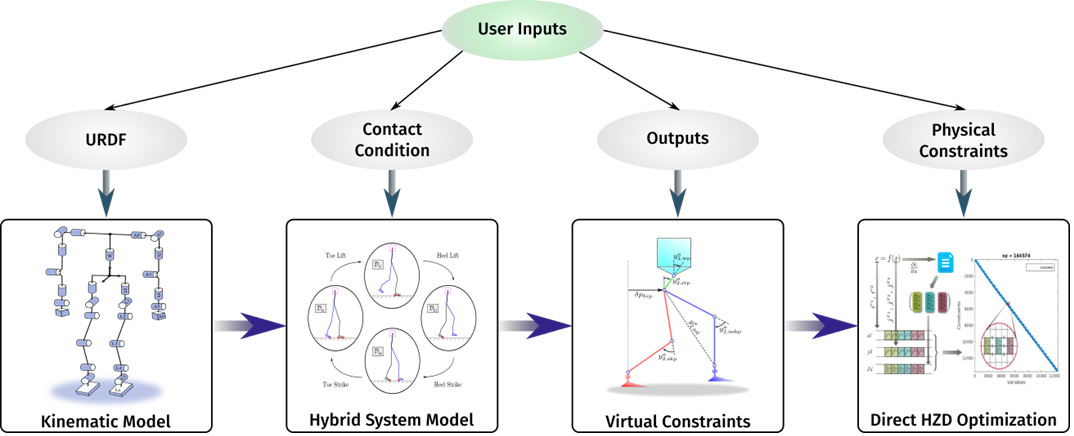}
  \caption{The process of modeling and optimizing dynamic walking gaits in FROST. Given a robot model (e.g., via a URDF file), FROST makes it easy for users to define the desired behavior and provide the necessary physical constraints to formulate a realistic gait optimization problem.}
  \label{fig:frost-process}
\end{figure*}

In a hybrid system model, each vertex represents an admissible continuous phase (a.k.a.  domain) determined by the differential equation of motion and physical constraints of the robot. These constraints include the contacts or mechanical constraints such as four-bar links or contacts with the environment, which modeled as \emph{holonomic constraints}, and associated limiting conditions such as the friction cone or swing foot clearance conditions, which can be formulated as a set of \emph{unilateral constraints}. A \emph{holonomic constraint} establish a set of algebraic equations to the continuous system dynamics. Moreover, for each edge in the graph, a \emph{unilateral constraint} determines the guard condition that triggers the associated discrete event. For example, in legged locomotion, the robot switches to different continuous domains when it establishes new contacts or breaks existing contacts. FROST describes the discrete behaviors of a robot by defining these kinematic and dynamic elements of the robot in the framework of a hybrid system model given in \eqref{eq:hybrid-control-system}.

\newsec{Continuous Dynamics.} The continuous domain describe the evolution of system states governed by differential algebraic equations (DAEs) on a smooth manifold. Let $x\in\mathcal{Q}$ be a set of coordinates of the system, the continuous dynamics is governed by the following ordinary differential equations (ODEs):
% \begin{align}
%   \label{eq:continuous-dynamics}
%   \tag{Second Order ODEs}
%   M(x)\ddot{x} = F(x,\dot{x}) + G(x,u)
% \end{align}
\begin{align}
  \label{eq:second-ode}
  \quad M(x)\ddot{x} = F(x,\dot{x}) + G(x,u)
\end{align}
subject to a set of algebaric equations $h(x) = 0$, where $M(x)$ is the positive definite mass matrix, $F(x,\dot{x})$ is a set of drift vectors, and $G(x,u)$ is a set of input vectors with $u$ being the system inputs. FROST also allows to use a first-order ODEs to describe the system equations of motion \cite{Hereid2017FROST, chen2017enhancing}.

\newsec{Discrete Dynamics.} When the evolution of a system transitions from one continuous phase into another, its states often undergo an instantaneous discrete change. It can be given expressed in the following manner:
\begin{align}
  \label{eq:reset-map-x}
  x^+ = \Delta_{x}(x^-),
\end{align}
where $x^- = \lim_{t \nearrow t_0} x(t)$ and $x^+ = \lim_{t \searrow t_0} x(t)$ with $t_0$ be the time instant at which the discrete dynamics occurs, and $\Delta_{x}$ represents a reset map at transition. For a second-order system, the first order derivative of $x$ should also be considered, i.e.,
\begin{align}
  \label{eq:reset-map-dx}
  \dot{x}^+ = \Delta_{\dot{x}}(x^-)\dot{x}^-, 
\end{align}
where $\dot{x}^- = \lim_{t \nearrow t_0} \dot{x}(t)$ and $\dot{x}^+ = \lim_{t \searrow t_0} \dot{x}(t)$.

\begin{remark}
  Given a robot's kinematic structure and inertial parameters, FROST provides many useful functions to easily compute the forward kinematics and dynamics of the system \cite{Hereid2017FROST}. A typical process in FROST for the modeling and gait design of a bipedal robot is shown in \figref{fig:frost-process}.
\end{remark}

\subsection{Trajectory Optimization}

The key functionality of FROST is to provide a fast and scalable trajectory planning algorithm for the hybrid dynamical systems. Given a hybrid dynamical system model in \eqref{eq:hybrid-control-system}, FROST automatically constructs a multi-phase hybrid trajectory optimization problem \cite{Hereid2016Thesis}. In particular, it uses direct collocation methods to solve such a trajectory optimization problem. FROST supports various types of direct collocation methods, including local collocation methods that use Hermite-Simpson or Trapezoidal schemes and global collocation methods that collocate at Legendre-Gauss-Lobatto (LGL) points. For the mathematical formulation of different collocation methods, we refer \cite{Hereid2016Thesis} to the readers for more detail. The result of such a trajectory optimization problem is a large-scale sparse nonlinear programming (NLP) problem. In general, it can be stated as:
\begin{align}
  \label{eq:gait-opt-frost}
  \hspace{-2px}\argmin  &\sum_{j=1}^P\left(\sum_{i=1}^{N_j}{w_i \mathcal{L}_j(x_i,\dot{x}_i,u_i)   + E_j(x_0,u_0,x_{N_j},u_{N_j})}\right) \\
  \hspace{-5px}\st &\tag{dynamics} M_j(x_i)\ddot{x}_i = F_j(x_i,\dot{x}_i) + G_j(x_i,u_i), \\
                        &\tag{collocation} \delta_j(x_0,\dots,x_i,\dot{x}_i,\ddot{x}_i,\dots,\ddot{x}_{N_j}) = 0,\\
                        &\tag{reset map} (x^{j+1}_0,\dot{x}^{j+1}_0) = \Delta_{j}(x^{j}_{N_j},\dot{x}^{j}_{N_j}),\\
                        &\tag{path constraints} C_j(x_i,\dot{x}_i,u_i) \geq 0
\end{align}
where $\mathcal{L}_j(\cdot)$ represents the running cost and $E_j(\cdot)$ represents the terminal cost defined in the domain $j\in\Vertex$ which has $P$ vertices, $\delta$ is the collocation constraints, $\Delta_{j}$ is the reset map associated with domain $(j,j+1)\in\Edge$, and $C(\cdot)$ is the collection of physical constraints, such as foot clearance, joint angle/velocity limits and torque limits, etc.

%%% Local Variables:
%%% mode: latex
%%% TeX-master: "main"
%%% End:

\section{C-FROST Implementation}
\label{sec:cfrost}

% TODO Show alternate workflow using C-Frost Mention using: (1) Ubuntu (2) Intel MKL when using Blas (3) Mention what tested solver is used with Ipopt (4) Mention native C++ Ipopt

% Pseudo algorithm for computing cost/constraints and gradient/constraints Convert symbolic expressions

This section presents C-FROST\footnote{C-FROST is publically available at \url{https://github.com/UMich-BipedLab/C-Frost}.}, an open source package that allows the user to convert and run the trajectory optimization problem in a native C++ environment instead of MATLAB. By reducing the computational overhead in the MATLAB environment, C-FROST noticeably speeds up the computational performances. It also comes with a few added benefits such as:
\begin{itemize}
\item using multi-threading techniques to speed up the evaluation of function Jacobians; and
\item making it easier to execute multiple optimization problems in parallel on multi-core processors; and
\item allowing easier deployment of parallel optimization to remote cloud services.
\end{itemize}

\subsection{Formulation of NLP Problems}

The use of direct collocation converts all constraints as algebraic constraints. While in more traditional trajectory optimization or optimal control problem, the state trajectories must be obtained by integrating the differential equations of motion. This enables us to construct the optimization problem as a general purpose nonlinear programming (NLP) problem, stated as:
\begin{align}
  \label{eq:gait-opt}
  X^* = \underset{X}{\argmin} \quad &\sum_{i=1}^{N_o}\mathcal{L}_i(X_i) \\
  \st &\mathcal{C}^l_j \leq \mathcal{C}_j(X_j) \leq \mathcal{C}^u_j \quad \forall j \in [1,\dots,N_c],\nonumber \\
                                    &X^l \leq X \leq X^u, \nonumber
\end{align}
wher $X \in R^{N_x}$ with $N_x$ being the total number of optimization variables, $N_o$ is the total number of cost functions with each cost function $\mathcal{L}_i$ has dependent variables $X_i$, where $X_i$ is a collection of components of $X$, $N_c$ is the total number of constraint functions, each constraint $\mathcal{C}_j$ has dependent variables $X_j$ (again, $X_j$ is also a collection of components of $X$), and $\mathcal{C}^l_j$ and $\mathcal{C}^u_j$, and $X^l$ and $X^u$, respectively, are the lower and upper bounds of $\mathcal{C}_j$ and $X$. We also denote $\mathcal{C} = [\mathcal{C}_1;\dots;\mathcal{C}_{N_c}]$ as the vector of all constraints, and $\mathcal{C}^l$ and $\mathcal{C}^u$ as the lower and upper bound of $\mathcal{C}$, respectively.

A particular advantage of using FROST is its capability to compute the closed-form symbolic expressions of all constraints and cost functions and export them to optimized C++ codes for fast computation. This is realized by the custom designed MATLAB symbolic math toolbox in FROST powered by the Wolfram Mathematica kernel as the backend. While formulating the problem in \eqref{eq:gait-opt}, we introduce additional \emph{defect variables} to simplify the constraints expressions in FROST. That enables FROST to compute the analytic derivatives (a.k.a., gradient or Jacobian matrices) of constraint and cost functions with respect to the corresponding dependent variables, as well as to determine the non-zero entries in the derivatives. In C-FROST, we generate two functions to compute the sparse Jacobian matrices: one that returns the row and column indices of all non-zero entries, and another one computes the values of these non-zero entries. Using these two functions, one can formulate the sparse Jacobian matrix of a constraint or the gradient of the cost function.

\subsection{Execution of NLP Problems}

C-FROST contains two main components: a collection of MATLAB functions that convert and export the original optimization problem in MATLAB to a set of C++ functions and JSON configuration files; and a C++ program that uses these exported functions and configuration files to run the optimization on a native Linux environment by calling the IPOPT solver directly.

\newsec{Exporting the NLP Problem.} To export an NLP problem to the C-FROST format, we extract the structure of the problem in \eqref{eq:gait-opt} and export this information to a collection of JSON files. For each constraint, the following information is required:
\begin{itemize}
\item the associated C++ functions, $f_j$ to compute the constraint or cost function, $G_j$ to compute the non-zeros in the Jacobian matrix, and $Gs_j$ to return the indices of non-zero terms;
\item the indices of dependent variables $X_j$ in $X$;
\item the indices of constraints $\mathcal{C}_j$ in $\mathcal{C}$;
\item the auxiliary constant data \texttt{aux} required to compute $f_j$;
\item the indices of non-zero Jacobian entries $G_j$ in $G_{nz}$.
\end{itemize}
The above information is constructed as arrays of structured data, \texttt{Constraints}, and exported to a JSON file. The same information for the cost functions is also formulated and exported to the same JSON file. We also generate two additional configuration files with one storing the boundary values of constraints and optimization variables and another storing the initial guess for the NLP solver. The C++ functions $f_j$, $G_j$ and $Gs_j$ are generated automatically using the symbolic math toolbox in FROST.

% The different symbolic expressions denoting the various constraints and objective functions are converted to C++ functions along side their jacobians.

% In addition, a few JSON files are generated that contain various configurations and bounds for constraints, such as feet height during walking, and variables, such as joint limits and torques.

% The generated code, along side C-FROST provided code, are then compiled in an executable.
\begin{algorithm}[t]
  \caption{Evaluation of the NLP Constraints}
  \label{alg:func}
  \begin{algorithmic}[1]
    \Procedure{EvalFunc}{$X$, \texttt{Constraints}} \State Initialize $\mathcal{C} \gets 0$ \ForAll{$f_j \in $ \texttt{Constraints}} \State $x_j \gets$ the dependent variables of $f_j$ from $X$ \State \texttt{aux} $\gets$ the auxiliary constants of $f_j$ \State $\mathcal{C}_j \gets$ the evaluation of $f_j(x_j; $ \texttt{aux}$)$ \State $r_j \gets$ the indices of $\mathcal{C}_j$ values in $\mathcal{C}$ \State $\mathcal{C}(r_j) \gets \mathcal{C}_j$ \EndFor \State \textbf{return} $\mathcal{C}$ \EndProcedure
  \end{algorithmic}
\end{algorithm}

\begin{algorithm}[t]
  \caption{Evaluation of the sparse Jacobian matrix of NLP Constraints}
  \label{alg:jac}
  \begin{algorithmic}[1]
    \Procedure{EvalJac}{$X$, \texttt{Constraints}} \State Initialize $G_{nz} \gets 0$, $i_R \gets 0$, $ j_C\gets 0$ \ForAll{$f_j \in $ \texttt{Constraints}} \State $x_j \gets$ the dependent variables of $f_j$ from $X$ \State \texttt{aux} $\gets$ the auxiliary constants of $f_j$ \State $J$ $\gets$ the evaluation of $G_j(x_j;$ \texttt{aux}$)$ \State $(i_r, j_c)$ $\gets$ the evaluation of $Gs_j(x_j;$ \texttt{aux}$)$ \State $r_j \gets$ the indices of $J$ values in $G_{nz}$ \State $G_{nz}(r_j) \gets J_i$, $i_R(r_j) \gets i_r$, $j_R(r_j) \gets j_c$ \EndFor \State \textbf{return} sparse($G_{nz}, i_R, j_C$) \Comment{Construct the sparse matrix} \EndProcedure
  \end{algorithmic}
\end{algorithm}

\newsec{Evaluating the NLP Problem.} Using the generated C++ functions, C-FROST will create a C++ program that calls the interior-point-based NLP solver IPOPT as a shared object to solve the direct collocation optimization problem. The program will first read the exported configuration files to construct a problem for IPOPT, and then call the IPOPT optimization routine, which requires sub-routine functions that compute the cost function and its gradient, the constraints, and the sparse Jacobian of the constraints. In Procedure~\ref{alg:func}, we describe how the constraints are computed. The same procedure applies to the cost function. Procedure~\ref{alg:jac} illustrates the construction of the sparse Jacobian matrix for constraints, using the generated C++ functions and other indexing information. Again, the same procedure is used for calculating the gradient of the cost function. It is important to note that we will only need to compute the non-zero entries in the Jacobian matrix. Considering that the percentage of non-zero entries in the Jacobian of a FROST-generated NLP problem is often far less than $1\%$, this significantly reduces the overall computational load.

\subsection{Parallel Evaluation of NLP Problems}
\label{sec:parallel_evaluation}

To speed up the optimization further, we consider two types of parallelization in FROST. For a single gait optimization, C-FROST provides multi-thread parallelization for the evaluation of Jacobian, whereas to run multiple optimization problems together, we prefer to run them in parallel on a multiple CPU cores platform. 

\newsec{Subroutine Parallelization.} The evaluations of the Jacobian of some constraints might be computationally expensive, like in the case of the dynamics equation.
Therefore there are benefits to parallelization when evaluating such matrices. 
The subroutines of each constraint within the for-loop in Procedure~\ref{alg:jac} are independent of other constraints; hence they can be run in parallel.
C-Frost allows the different constraint Jacobians to be evaluated simultaneously by multi-threading Procedure~\ref{alg:jac}.
For each constraint $f_j$, lines 4 till 9 in Procedure~\ref{alg:jac} are evaluated in available workers.
To avoid concurrent manipulation of the shared data $G_{nx}$, $i_R$ and $j_C$, the evaluations in line 9 is locked using a monitor class to synchronize access to the shared data.
The evaluation of the constraint Jacobian is significantly more computationally expensive than the evaluations of the constraint functions, the cost functions and the gradient of the cost function.
And thus, these processes have not been parallelized.

\newsec{Parallel Optimization.} When having multiple optimization problems to be solved, as in the application of generating a library of gaits, running the multiple optimization problems simultaneously is preferred to using multiple threads per problem as that avoids the overhead from thread synchronization in a single program. This can be done by writing a simple bash script that starts new optimizations when a CPU is underutilized or existing third-party tools, such as the GNU Parallel package \cite{Tange2011a}.

The ability to run many different gait optimization in parallel is particularly useful for learning-based control approaches as described in \cite{da2017combining, da20162d}, and can also be used in the mechanical design process of a robot, where instead of gait characteristics varying, link lengths, mass distributions, and gear ratios are varied \cite{Westervelt2002Design, Reher2016Realizing}.
% The functions that computes the cost and constraint values are given in Algorithms~\ref{alg:func} and \ref{alg:jac}.

% SPARCITY AND JACOBIAN

% \subsection{Parallelization of NLP Problems}

% % TODO: parallel execuation of multiple problems
% % TODO: parallel computation of constraints, jacobians, etc.

% To speed up the computation further, C-FROST offers two approaches to utilize multi-thread parallelization on multiple CPU cores.

% \newsec{Subroutine Parallelization.}

% \newsec{Parallel Execution.} When attempting to run multiple optimizations in parallel, it is possible to write a simple startup script that starts new optimizations when a CPU is underutilized. Alternatively, results in this paper have used GNU Parallel \cite{GNUParallel} to do so.

% \begin{algorithm}
%   \caption{Startup script to run 1000 optimization in parallel when no less than 95 of the CPU is being utilized}
%   \label{alg:bash}
%   \begin{algorithmic}[1]
%     \State Initialize $N \gets 1000$ \State Initialize MAX\_CPU $\gets 95$ \ForAll{$i \in [0, N)$} \State Initialize PARAM\_FILE$\gets params\_i.json$ \State Initialize OUTPUT\_FILE$\gets output\_i.json$ \If{CPU\_USAGE $<$ MAX\_CPU} \State Start PROGRAM --output OUTPUT\_FILE --param PARAM\_FILE \EndIf \EndFor
%   \end{algorithmic}
% \end{algorithm}

%%% Local Variables:
%%% mode: latex
%%% TeX-master: "main"
%%% End:

\section{A Case Study: Cassie Blue}
\label{sec:results}

% TODO Mention what gaits are generated and that Corriolis term was removed Comparing single thread CFROST with Matlab FROST Show Dell, Surface, and Bruce Desktop speeds Show with/without Coriolis Show single thread vs multithread on Bruce's Desktop Show stick figure for one of the forward walking gaits
The efficiency of C-FROST will be illustrated by designing a library of different periodic walking gaits for the Cassie-series bipedal robot shown in \figref{fig:cassie}.

\subsection{Testbed Robot Model}
This underactuated robot's floating base model has 20 degrees of freedom (DOF). There are seven joints in each leg with five of them actuated by electric motors through gearing and the other two joints being passive where they are realized via a specially designed four-bar linkage with one bar being a leaf spring. When supported on one foot during walking, the robot is underactuated due to the narrow feet. The springs in the four-bar linkages also introduce additional underactuation. For simplicity, we assume these springs have infinite stiffness, i.e, rigid links, in this paper (see \figref{fig:cassie:coordinates}). The description of the mechanical model of Cassie used in this paper is publicly released on GitHub\footnote{Available at \url{https://github.com/UMich-BipedLab/Cassie_Model}.}.

In this paper, we will use FROST to model and formulate a series of periodic walking gait optimization problems for the Cassie, and then use C-FROST to generate these gaits in parallel rapidly. The optimization performance will be evaluated on multiple platforms---including remote cloud services---with different numbers of CPU cores. While the present paper is concerned with optimization for the open-loop model of Cassie, some of the designed gaits are implemented to demonstrate that they are physically realistic, and for that, a control policy is necessary. The control laws used in the simulations and experiments reported here are based on the work in \cite{gong2018feedback}.  To enhance our confidence that the presented optimization results are representative of what other users will find on their models, we have also run several of them on an 18-DOF exoskeleton model presented in \cite{Harib2018Feedback}. These results are not reported here.

\subsection{Optimization Problem}
\label{sec:cassie-opt}

The primary problem being illustrated in this paper is to design two-step periodic walking gaits for Cassie Blue. This walking pattern is modeled as a two-domain (a right-stance domain and a left-stance domain) cyclic hybrid system model in FROST. A virtual constraints based feedback controller is also enforced in the gait optimization problem so that the results are not just the open-loop join trajectories but also a class of feedback controllers that can be used to stabilize the periodic motion in the full-order robot dynamics \cite{Hereid2016Thesis}.

The optimization problems are formulated in FROST using the structure described in \eqref{eq:gait-opt}. When formulating the gait optimization problems, we impose periodicity over two steps for a desired average walking speed and ground inclination. The cost function to be minimized in the optimization is the ``pseudo energy'', given as
\begin{align}
  \label{eqn:costFun}
  \mathcal{L}_j = \|u_i(t)\|^2,
\end{align}
where $\mathcal{L}_j$ is the integral function for the running cost defined at the domain $j$, with $j\in\{1,2\}$ being the index of the right and left stance domain and $u_i$ being the control inputs at each collocation node. The ``pseudo energy'' in \eqref{eqn:costFun} is not normalized by step length because the speed and step duration are constrained. In addition, the optimization also captures torque and joint limits of the Cassie robot as well as the following physical constraints that are heuristically chosen to achieve stable walking. These constraints are:
\begin{itemize}
\item[$\bullet$] Fixed step duration $T$ of $0.4$ s;
\item[$\bullet$] Swing foot clearance of at least $15$ cm at midstep;
\item[$\bullet$] Ground reaction forces respect the friction cone and ZMP condition~\cite{Grizzle2014Models};
\item[$\bullet$] Zero swing foot horizontal speeds at impact;
\item[$\bullet$] For purely sagittal plane walking, require the two steps in a gait to be symmetric;
\item[$\bullet$] Sagittal plane walking gaits have tighter bounds on hip abduction and rotation motors than gaits that include lateral motion; and
\item[$\bullet$] Torso must remain upright within $\pm 3$ degrees pitch and roll limits.
\end{itemize}

\begin{figure}
  \vspace{2mm} \centering
  \begin{subfigure}[t]{0.235\textwidth}
    \centering \includegraphics[width=1\columnwidth]{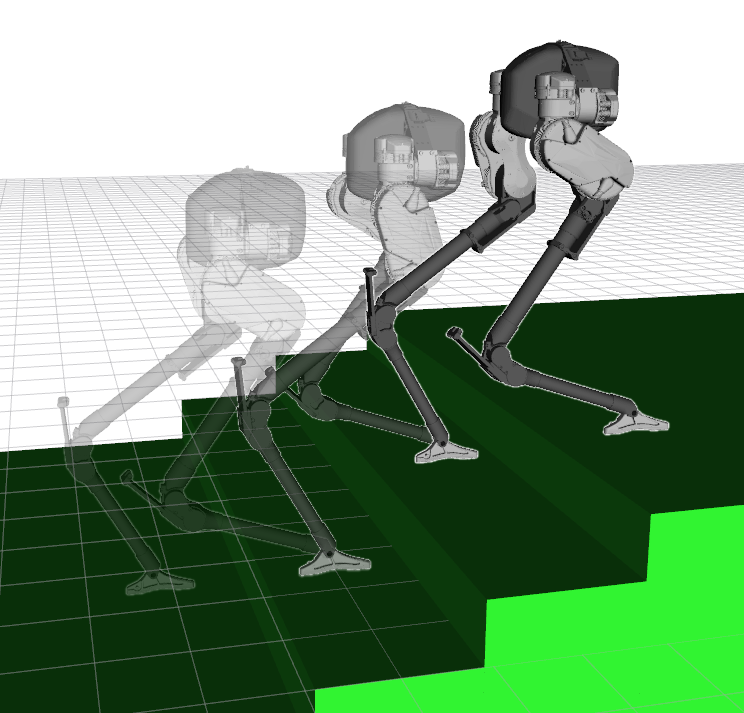}
    \caption{$15$ cm step up while walking forward at $1$ m/s.}
    \label{fig:gaits:step-up}
  \end{subfigure}
  \begin{subfigure}[t]{0.235\textwidth}
    \centering \includegraphics[width=1\columnwidth]{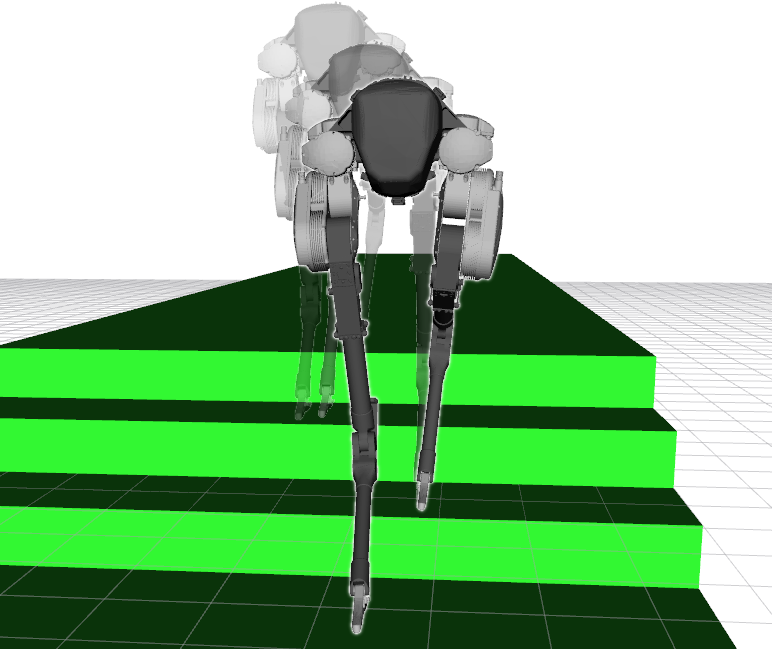}
    \caption{$15$ cm step down while walking forward at $1$ m/s and left at $0.3$ m/s.}
    \label{fig:gaits:step-down}
  \end{subfigure}
  \caption{Stereoscopic tiles for stepping up/down at different walking speeds.}
  \label{fig:gaits}
\end{figure}

In particular, we use C-FROST to design a library of $11^3 = 1,331$ periodic gaits defined on a cubic uniform grid with different walking speeds and ground inclinations:
\begin{itemize}
\item[$\bullet$] 11 different backward/forward walking speeds ranging from $-1$ m/s to $1$ m/s;
\item[$\bullet$] 11 different left/right walking speeds ranging from $-0.3$ m/s to $0.3$ m/s; and
\item[$\bullet$] 11 different inclinations of the ground surface from $-15$ cm to $15$ cm per step.
\end{itemize} 
The desired average walking speed is enforced via an equality constraint
\begin{align}
    \bar{v}_x = \frac{p^x_N - p^x_0}{T}, \quad
    \bar{v}_y = \frac{p^y_N - p^y_0}{T}
\end{align}
where $p^x_N$, $p^x_0$, $p^y_N$, and $p^y_0$ are the (X,Y) position of the hip at the end and beginning of a step, respectively, and $\bar{v}_x$ and $\bar{v}_y$ are the desired sagittal and lateral walking speeds. The desired ground inclination will be enforced by changing the height where the swing foot should impact with the ground. Two of the gaits from this set are illustrated in \figref{fig:gaits}.

\section{Results}
\label{sec:experiments}
This section first presents the computational performance for creating a collection of periodic walking gaits for Cassie Blue under different scenarios using C-FROST\footnote{The source code of this work is now publically available on \url{https://github.com/UMich-BipedLab/Cassie_CFROST}.}. Then we validate the feasibility of the optimized walking gaits on Cassie through experiments.

\subsection{Computational Performance Analysis}
\label{sec:results:cfrost-opt-performance}

To show how C-FROST can be used to improve the gait optimization process, we first compare its performance with the original MATLAB implementation in FROST, and then secondly, we show the improvement in performance when using parallelization on different platforms, including commercial cloud servers with large numbers of CPU cores. 

% Note that, we specifically remove the velocity terms in the rest of the tests.

% In all cases, the same compiled binaries are used.  From the table, one sees that, for a large batch of gaits, on an i7-class processor, \textbf{Only works for Surface}
% \begin{equation}
%     \label{eq:TotalTime}
% \text{Total Optimization Time}\approx \frac{N_{\rm gaits}}{N_{\rm threads}}6.2~~\text{seconds}, 
% \end{equation}
% assuming two threads per core.

\newsec{Comparison versus FROST.}  In this comparison, we tested running a single gait optimization using FROST on MATLAB and using C-FROST running as a native program on Ubuntu. The optimization is set to design a walk-in-place gait (zero sagittal and lateral speeds) on flat ground. Both tests were run on a Thinkpad laptop with a 2.6GHz Intel i7 processor, under identical settings, and with the same initial guess. No parallelization was enabled in this test.

In our test, it took $90.12$ seconds to generate the walk-in-place gait when running on MATLAB, whereas, using C-FROST resulted in the gaits being generated in $24.51$ seconds. The result shows that using C-FROST speeds up the gait optimization process by approximately $3.5$ times for a single gait. This demonstrates that C-FROST can be used to rapidly generate dynamic motions for legged robots without dealing with the computational overhead associated with MATLAB while still allowing one to use the easy-to-prototype environment provided by FROST. It should be noted that the comparison is the time it takes to run the optimizations and excludes the setup time to define the optimization problems since the latter is essentially the same for both methods.

% \begin{table*}[t]
%   \vspace{3mm} \centering
%   \caption{Time to perform $11^3=1,331$ optimizations for the Cassie URDF model, without the velocity terms. The same pre-compiled code is run on all platforms. The average compute time for a gait is similar in all cases; the total times essentially inversely scale with the number of threads with slight differences with regards to the CPU frequency.}
%   \vspace*{0.1cm}
%   \label{tab:pc_opt_times}
%   \begin{tabular}{ c | c | c | c}
%     Computing Resource & CPU Cores & OS & Time (min)  \\ \hline \hline 
%     Surface Pro 2017  i7-7660U @ 2.5GHz & 2 cores (4 threads) & Ubuntu 16.04 on WSL & 67.09 \\ 
%     Dell XPS 9560 i7-7700HQ @ 2.8GHz & 4 cores (8 threads) & Ubuntu 16.04 n WSL & 38.87 \\
%     Dell XPS 9560 7-7700HQ @ 2.8GHz & 4 cores (8 threads) & Ubuntu 16.04 & 36.88 \\
%     Desktop 9-7900X @ 3.30GHz  & 10 cores (20 threads) & Ubuntu 16.04 & 13.83 \\ %12.96 with new bounds
%     Google GCP Xeon Skylake @ 2.0 GHz & 24 vCPUs (24 threads) & Ubuntu 16.04 & 16.95 \\
%     Amazon EC2 Xeon E5-2666 v3 & 36 vCPUs (36 threads)& AMI Linux 4.14 & 8.85 \\
%     Amazon EC2 Xeon Platinum @ 3.0 GHz & 72 vCPUs (72 threads)& AMI Linux 4.14 & 4.32 \\ \hline
%   \end{tabular}
% \end{table*}

\begin{table*}[t]
  \vspace{3mm} \centering
  \caption{Time to perform $11^3=1,331$ gait optimizations for the Cassie. The same pre-compiled code is run on all platforms. The average compute time for a gait is similar in all cases; the total times essentially inversely scale with the number of threads with slight differences with regards to the CPU frequency.}
  \vspace*{0.1cm}
  \label{tab:pc_opt_times}
  \begin{tabular}{ c | c | c | c}
    Computing Resource & CPU Cores & OS & Time (min)  \\ \hline \hline 
    Desktop 9-7900X @ 3.30GHz  & 10 cores (20 threads) & Ubuntu 18.04 & 401.18 \\ %12.96 with new bounds
    Google GCP Xeon Skylake @ 2.0 GHz & 64 vCPUs (64 threads) & Ubuntu 18.04 & 113.93 \\
    Amazon EC2 Xeon Platinum @ 3.0 GHz & 72 vCPUs (72 threads)& AMI Linux 4.14 & 91.11 \\ \hline
  \end{tabular}
\end{table*}

\begin{figure}
  \vspace{2mm} \centering \includegraphics[width=0.45\textwidth]{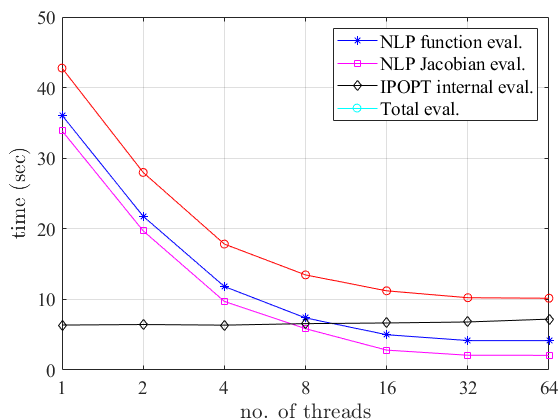}
  \caption{The computation time of each components while using different number of threads.}
  \label{fig:jacobian-parallel}
\end{figure}

\newsec{Single gait optimization with subroutine parallelization.} Further analyzing the result of C-FROST optimization, approximately $80\%$ of the total computational time was spent to evaluate the NLP functions and the rest was used for IPOPT internal computation. Moreover, approximately $85\%$ of the NLP function evaluation time was used to compute the sparse Jacobian matrix of the constraints. It indicates that computing Jacobian matrix becomes the main performance bottle neck. 

Here, we report the performance improvements while enabling multi-threading for the computation of Jacobian matrix as described in \secref{sec:parallel_evaluation}. In this test, we ran the walk-in-place gait optimization on a Linux virtual machine on the Google Cloud Platform. The results are shown in \figref{fig:jacobian-parallel}. The plot shows that while the time required for IPOPT internal computation and other user-defined NLP functions other than the constraint Jacobian (the different between blue and purple line) remaining the same, the time to compute the constraint Jacobian (the purple line) reduces significantly when the available number of threads increases. It reduces from $34$ (single core) to $2$ seconds (64 cores), which yields a $17$ times speed-up. When having more than 16 cores, however, the overhead time for synchronization between threads will dominate the overall run time, and the change is barely noticeable. Moreover, the affect on the total evaluation time becomes less when it requires more time for IPOPT internal computation. Overall, the total optimization evaluation sped up from $42.8$ seconds to $10.2$ seconds when using up to 64 cores.

\newsec{Generating multiple gaits in parallel.} The speed-up factor of the subroutine parallelization is not significant enough if considering the number of CPU cores being used. In need of generating multiple gaits at the same time, running all optimizations in parallel on available CPU cores will be more efficient. In our final test, all $1,331$ gaits were generated in a single batch by running the optimization solvers in parallel using all available cores on various computing platforms. The JSON configuration files that specify the boundary condition for each optimization are generated a priori, and then multiple instances of the same compiled program were set to run on different CPU cores using different boundary conditions under identical settings. When attempting to run multiple optimizations in parallel, we use the GNU Parallel package. ~\tabref{tab:pc_opt_times} shows the computation times, specific machines, number of cores, and OS. The results show that utilizing all available computing resources of the given platforms, C-FROST can dramatically reduce the total time to compute batches of gaits. This feature is particularly important for learning based walking controllers, which require that one generate a large set of gaits as training data. In particular, users can take advantage of commercial cloud servers, such as Amazon's AWS or Google's GCP to generate a large set of gaits in a much shorter period. For example, using 72 vCPUs on AWS yields an average-per-gait computation time of approximately $4.1$ seconds. This result shows that C-FROST noticeably reduces the computation time of trajectory optimization for a high-dimensional hybrid dynamical system without compromising problem formulation with simplified models.

% An interesting observation is that run times on native Ubuntu vs using Ubuntu on the Windows Subsystem for Linux (WSL) are similar.

% TODO Show some plots/figures of walking Cassie robot Speed tracking?  Basic controller explanation? (or cite MARLO paper?)

\subsection{Experimental Validation}

The purpose of the experiments is to assure the reader that the set of constraints imposed during trajectory optimization are physically realistic and were not selected to simplify the designs. The controller that realizes the trajectories on Cassie Blue is based on \cite{da20162d} as detailed in \cite{gong2018feedback}. Note that, only a subset of gaits from the previously generated 1,331 gaits are being utilized in the experiment. Regulating lateral velocity and ground inclination that uses the full set of gaits will be subject to future work.

\begin{figure}[t]
  \vspace{2mm} \centering
  \begin{subfigure}[t]{0.237\textwidth}
    \centering \includegraphics[width=1\textwidth]{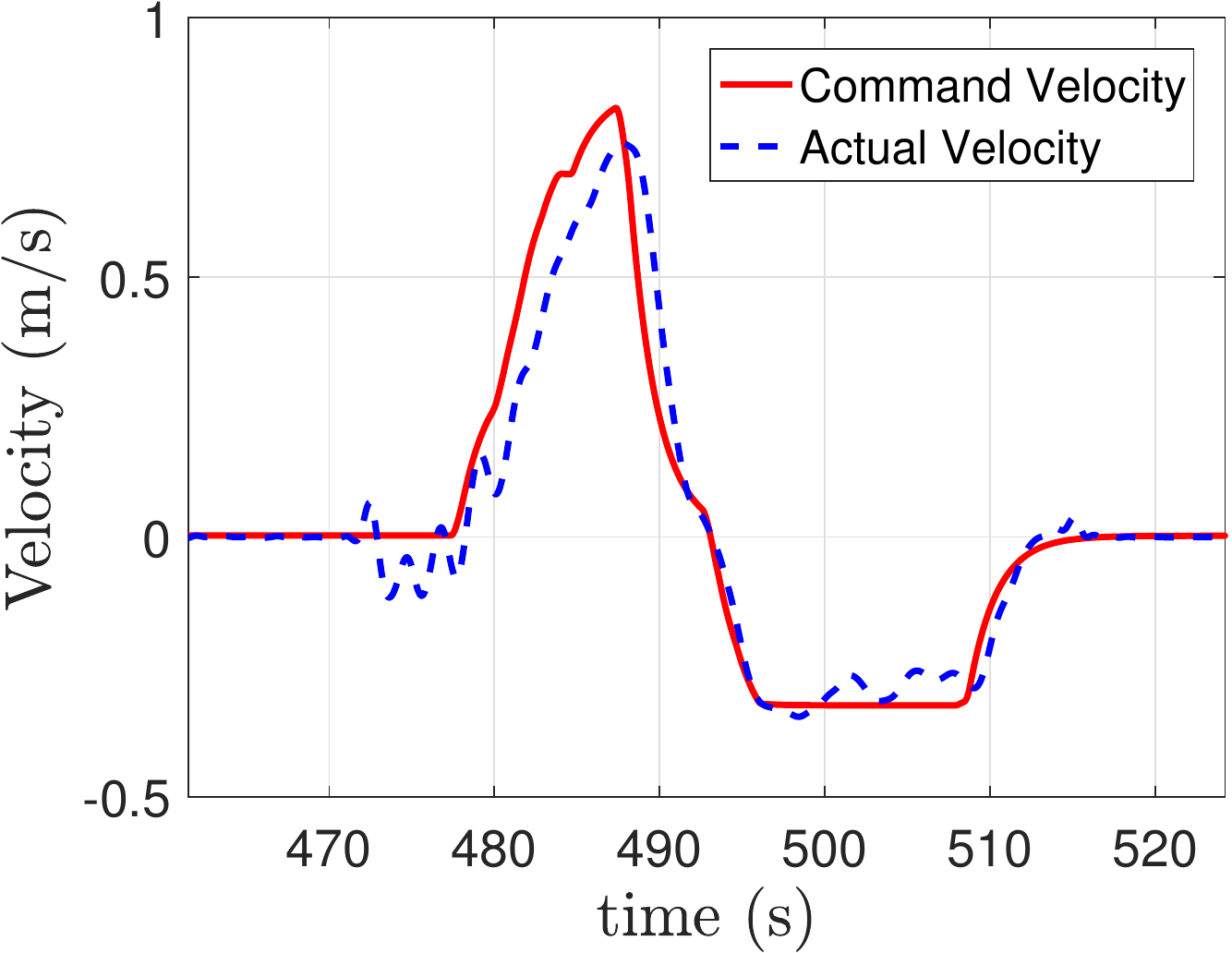}
    \caption{The velocity tracking performance of Cassie walking.}
    \label{fig:vel_track}
  \end{subfigure}
  \begin{subfigure}[t]{0.23\textwidth}
    \centering \includegraphics[width=1\textwidth]{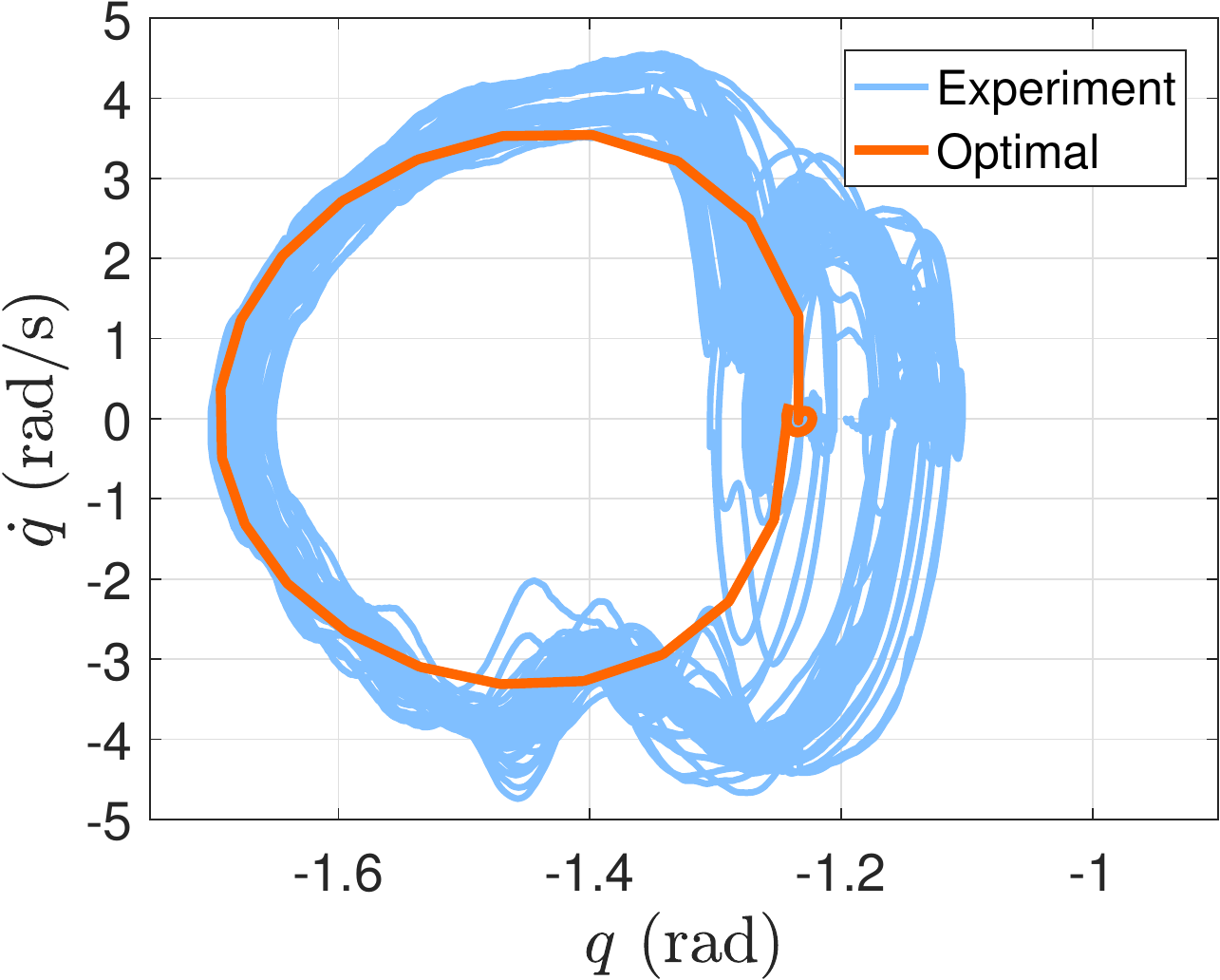}
    \caption{The phase portrait of the right knee joint during stepping in place.}
    \label{fig:knee_orbit_exp}
  \end{subfigure}
  \caption{Experimental results for a gait-library-based controller on Cassie Blue.}
  \label{fig:exp}
\end{figure}

During these tests, the $11$ sagittal gaits with speeds ranging from -$1.0$ m/s to $1.0$ m/s on flat ground were used to generate a gait library \cite{da20162d,gong2018feedback} for Cassie Blue to regulate the forward and backward walking speeds. A heuristic foot placement controller was also applied to stabilize the lateral motion of the robot. In the experiment, we use a remote controller to command a reference speed profile, and the robot utilized the gait library to follow the commanded speeds. The robot can accelerate from $0$ m/s to $0.8$ m/s in around $10$ seconds and then slow down to walking in place (e.g., $0$ m/s) in $5$ seconds. The robot is then commanded to walk backward with a nearly constant velocity of $0.3$ m/s across the room to return to the starting point. \figref{fig:vel_track} shows the tracking of the commanded velocity in the experiment. Also, \figref{fig:knee_orbit_exp} compares the similarity in the phase portraits of the right knee angle from optimization and experiment when the robot is stepping in place (i.e., zero forward velocity). A video that illustrates the different walking motions from the gait library built from optimization and the presented experimental results is available at \cite{CassieExpVideo}. %In the future work, we plan to implement the library of $1,331$ gaits on Cassie Blue to achieve versatile uneven terrain walking.

\section{Conclusions}
\label{sec:conclusion}
% \vspace{-3mm}

% The dynamic models of today's bipedal robots are hugely complex. This has led to some research groups advocating the use of simplified models while other groups, such as ours, have steadfastly used the full dynamic model when doing control design. This paper has shown that the ``full'' models can be simplified without altering the kinematic relations and with seemingly minimal consequence on gait designs for a Cassie-series biped walking at 1 m/s. The ``trick'', well known to the manipulator community, is to remove the ``velocity terms''. Here it allowed gait optimization times to be reduced by approximately 36 times on average and reduced model compile times in C++ by 90\%.

The primary objective of the paper was the presentation of a method to speed up trajectory design for high-dimensional systems by 3-4 times faster for single trajectories if not using parallelization and up to 9 times faster with parallelization. Moreover, the average computational time for a gait reduced significantly for families of trajectories. This was accomplished through the design of C-FROST, an open-source C++ interface for FROST---a MATLAB-based direct trajectory optimization tool for robotic systems. The significant speed up was illustrated by the design of a library of walking gaits for a Cassie bipedal robot. For a single gait, C-FROST completed an optimization in $24.51$ seconds vs $90.12$ seconds in FROST. The speedup is much more significant when doing multiple gaits, such as computing a (periodic) gait library \cite{da20162d} or building an invariant surface of open-loop transition trajectories \cite{da2017combining}. The paper documents the parallel computation in C-FROST of $1,331$ gaits for a Cassie-series bipedal robot on a range of platforms. The average computation time is roughly 4.1 seconds per gait while running on an AWS virtual machine. In future work, we will consider using other parallel platforms, such as GPUs, to speed up the trajectory optimization problems further.

% Currently, the time to run an optimization is essentially equal to the problem set up time, which in FROST, takes about a second. This paper has not looked at reducing that part of the design process, which will be the subject of future work.

%%% Local Variables:
%%% mode: latex
%%% TeX-master: "main"
%%% End:

%\section*{APPENDIX}

\section*{ACKNOWLEDGMENT} 
% The authors thank D. Da for suggesting that we look into the importance of the ``velocity terms'' in biped models.
The work of A. Hereid, O. Harib, and Y. Gong was supported by NSF Grant NRI-1525006.  The work of Y. Gong and R. Hartley is supported by the Toyota Research Institute (TRI) under award number N021515. The work of J. Grizzle was supported in part by NSF Grant NRI-1525006 and in part by TRI-N021515. The paper reflects only the opinions of its authors and not those of their funding bodies.

\bibliographystyle{plain}
\bibliography{bibliography/control,bibliography/hzd,bibliography/optimization,bibliography/robotics,bibliography/miscellaneous}

\end{document}